\documentclass{article}

\usepackage{microtype}
\usepackage{graphicx}
\usepackage{subfigure}
\usepackage{booktabs}
\usepackage{multirow}
\usepackage{makecell}
\usepackage{hyperref}
\usepackage{enumitem}
\usepackage{mathrsfs}
\usepackage{float}
\usepackage{ragged2e}

\usepackage[accepted]{icml2020_internal}

\icmltitlerunning{Aligning the Pretraining and Finetuning Objectives of Language Models}

\begin{document}

\twocolumn[
\icmltitle{Aligning the Pretraining and Finetuning Objectives of Language Models}

\begin{icmlauthorlist}
\large
\textbf{Nuo Wang Pierse} \quad \textbf{Jingwen Lu}
\vskip 0.1in
Bing Core Relevance, Microsoft
\end{icmlauthorlist}

\icmlaffiliation{bing}{Bing Core Relevance, Microsoft}
\icmlaffiliation{corr}{Bing Core Relevance, Microsoft}

\icmlcorrespondingauthor{Nuo Wang Pierse}{Nuo.WangPierse@microsoft.com}

\icmlkeywords{language models, transfer learning, deep learning applications}

\vskip 0.3in
]

\printAffiliationsAndNotice{}

\begin{abstract}
We demonstrate that explicitly aligning the pretraining objectives to the finetuning objectives in language model training significantly improves the finetuning task performance and reduces the minimum amount of finetuning examples required. The performance margin gained from objective alignment allows us to build language models with smaller sizes for tasks with less available training data. We provide empirical evidence of these claims by applying objective alignment to concept-of-interest tagging and acronym detection tasks. We found that, with objective alignment, our 768 by 3 and 512 by 3 Transformer language models can reach accuracy of 83.9\%/82.5\% for concept-of-interest tagging and 73.8\%/70.2\% for acronym detection using only 200 finetuning examples per task, outperforming the 768 by 3 model pretrained without objective alignment by +4.8\%/+3.4\% and +9.9\%/+6.3\%. We name finetuning small language models in the presence of hundreds of training examples or less ``Few Example learning". In practice, Few Example Learning enabled by objective alignment not only saves human labeling costs, but also makes it possible to leverage language models in more real-time applications.
\end{abstract}

\section{Introduction}

In the past three years, new deep learning language model architectures and techniques have significantly advanced the field of computational linguistics. The current state-of-the-art language models use large Transformer architectures \cite{Vaswani2017} containing hundreds of millions of parameters and are pretrained using multi-billion word corpuses. GPT \cite{Radford2018} and BERT \cite{Devlin2019} are two of the first examples of these Transformer-base language models. They outperformed the RNN-based models such as ELMo \cite{Peters2018} by a significant margin in benchmarks like GLUE \cite{Wang2019} and SQuAD \cite{Rajpurkar2016}. Since the release of GPT and BERT, many researchers have further improved the Transformer-based language models demonstrated by surpassing their predecessors in the evaluations of the common benchmarks \cite{Liu2019, Dong2019, Conneau2019, Conneau20192, Raffel2019, Yang2019, Sun2020}.

These new language model releases typically use more model parameters than their predecessors and are evaluated only on the academic datasets. This contrasts the two major challenges that we face in building natural language understanding (NLU) applications: (1) Speeding up language model inference; (2) Shortage of finetuning data for application-specific tasks. In our experience, the inference speed of the 768 by 12 BERT-base model, the smallest BERT model, is far from meeting the requirements for most real-time applications. To speed up language models, people have developed libraries for fast neural network computation \cite{Zhang2018, Junczysdowmunt2018} and built smaller models with sufficient prediction power using knowledge distillation \cite{Jiao2019} or improved Transformer architectures \cite{Lan2020}. In terms of the learning techniques for low-resource tasks, knowledge transfer through pretraining itself is a partial solution. GPT-2 has demonstrated its impressive zero-shot learning capability in text generation and question \& answering \cite{Radford2019}. Besides transfer learning, multitasking is another common technique used to boost low-resource task performance \cite{Lin2018, Liu2018, Conneau2019}.

In this paper, we tackle both of the above challenges at once. We develop a solution to train high-quality small language models using only a few hundred finetuning examples. We attempted to maximize the efficacy of knowledge transfer by designing pretraining objectives that closely resemble the finetuning objectives - we call this ``explicit objective alignment", ``objective alignment" in short. In our tasks, objective alignment not only enabled smaller language models to perform the finetuning tasks equally well as their larger counterparts pretrained without objective alignment, but also reduced the number of finetuning examples required. We were able to develop a concept-of-interest tagger and an acronym detector using a 768 by 3 Transformer model and 200 finetuning examples for each task. We call finetuning in this model size and data size limit Few Example Learning (FEL).

The main contributions of this paper are:
\begin{itemize}[topsep=0pt]
\item We propose pretraining objective alignment as a solution to developing small language models for NLU applications that have limited labeled data.
\item We demonstrate our solution by building two NLU applications of reasonable accuracy using a 768 by 3 Transformer model and 200 finetuning examples each.
\item We provide detailed steps to carry out object alignment, complete descriptions of the training parameters and recommendations for best practices.
\end{itemize}

\section{Objective Alignment and FEL}

Inspired by the success of zero-shot learning \cite{Radford2019}, we developed a solution - objective alignment - to tackle the problem of finetuning data scarcity. Objective alignment in the scope of this paper is a type of transfer learning that specializes in transferring the knowledge needed for the finetuning tasks of interest. ``Alignment" describes our attempt to design new pretraining objectives that resemble the finetuning objectives better than the standard pretraining objectives such as masked language model (MLM) and next-sentence prediction (NSP) \cite{Devlin2019}. We provide detailed examples of objective alignment in Section \ref{subsectionalignedobjectives}.

\begin{figure}[ht]
\begin{center}
\centerline{\includegraphics[width=6.5cm]{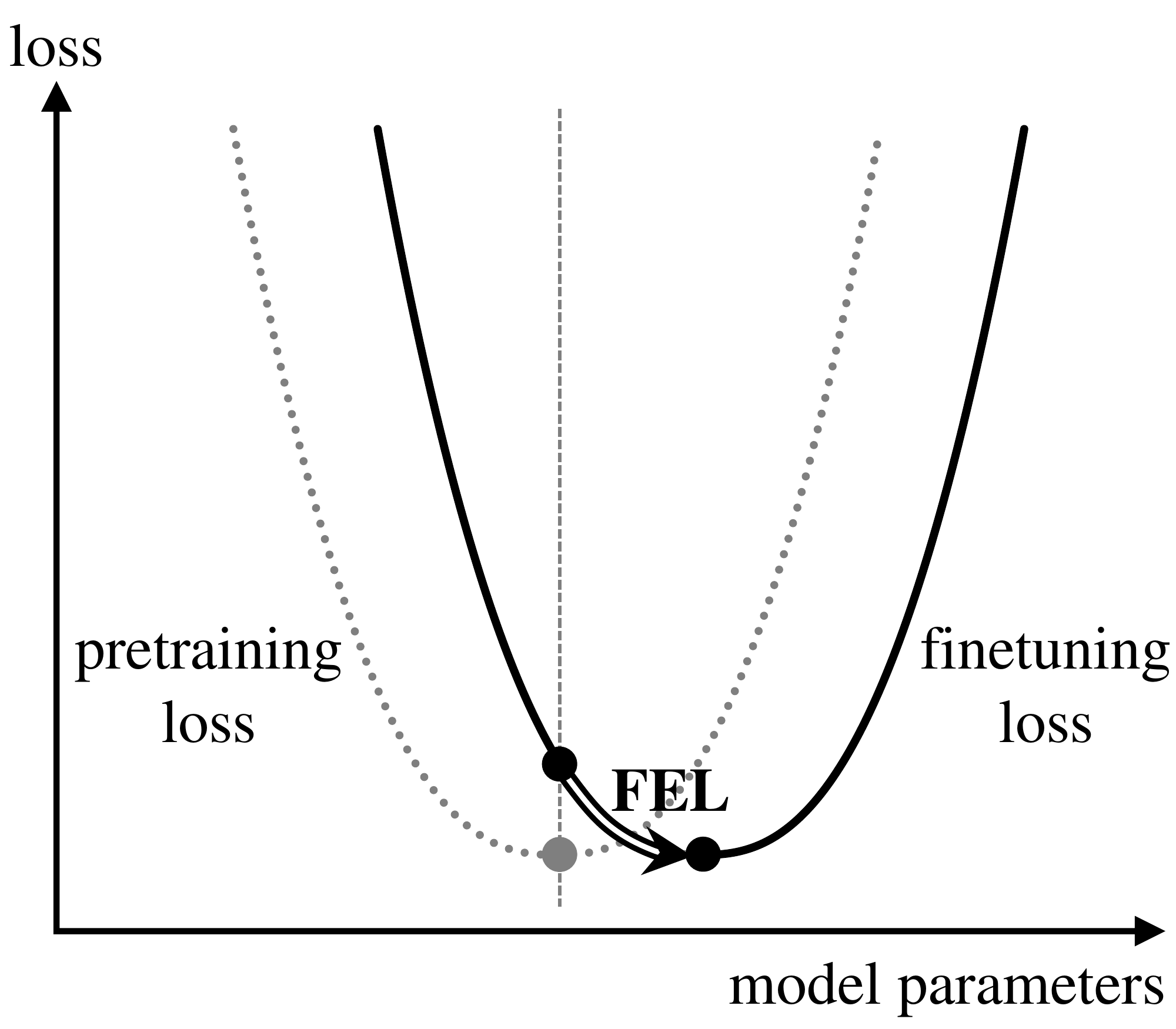}}
\caption{\textbf{A possible explanation of objective alignment and Few Example Learning (FEL).} The loss function minimum of an aligned pretraining objective (gray dot) is designed to be located near the loss function minimum of the finetuning objective (lower black dot) in the finetuning task model parameter subspace (``model parameters" in short). When the model switches from pretraining to finetuning, only a small amount of learning is necessary for finetuning convergence (double-dashed arrow).}
\label{figure1}
\end{center}
\vskip -0.2in
\end{figure}

FEL is the process of finetuning small language models using hundreds of examples or less after objective-aligned pretraining. We provide a possible explanation of FEL in Figure \ref{figure1}. During pretraining, the model undergoes multitask learning and converges along the additively combined loss functions of the standard pretraining objectives and the aligned pretraining objectives. The aligned objectives help the model converge closer towards the finetuning objective loss function minimum in the finetuning task model parameter subspace. And as a result, the model only needs a small amount of learning to converge in the finetuning phase. In our tasks, objective alignment and FEL allow a developer to produce the necessary amount of application-specific finetuning examples within hours and to finetune the model within minutes.

In this paper, we use the vanilla BERT architecture and focus our discussion on the independent contribution of objective alignment in improving language model performance. However, to build the most competent small language model, we recommend using objective alignment in combination with knowledge distillation \cite{Hinton2015, Liu22019, Jiao2019}, better Transformer architectures \cite{Dai2019, Yang2019, Lan2020}, better pretraining setup and pretraining data \cite{Liu2019, Dong2019, Conneau2019, Conneau20192, Raffel2019}.

\section{Methods}

\subsection{Finetuning Tasks}
\label{subsectiontasks}

\begin{figure}[ht]
\begin{center}
\centerline{\includegraphics[width=8.4cm]{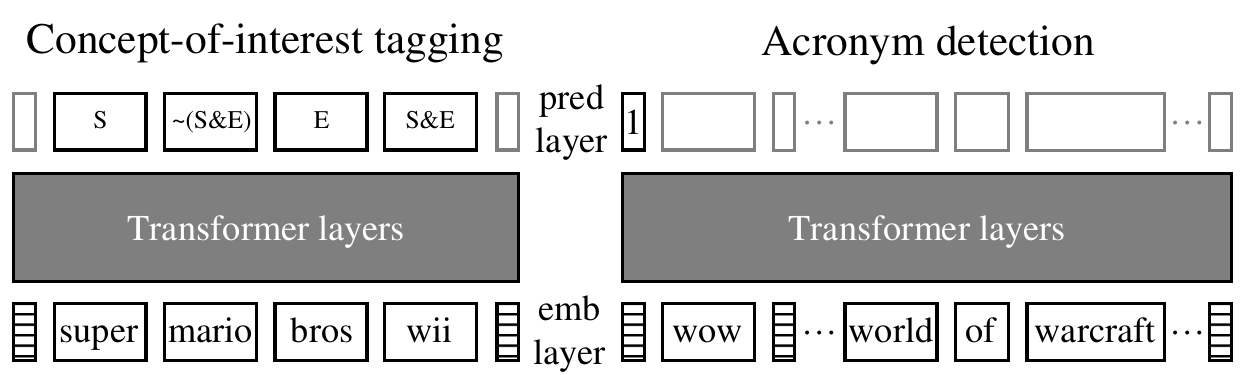}}
\caption{\textbf{BERT-like implementation of concept-of-interest tagging (CT) and acronym detection (AD).} CT is a token-level 4-class classification task. The four classes are: The start of a concept (S), the end of a concept (E), both start and end (S\&E), neither start nor end ($\sim$(S\&E)). CT uses ``[CLS] query [SEP]" input format. AD is a sentence-level binary classification task. The two classes are: acronym matches any entity in the snippet (1) and no match (0). AD uses ``[CLS] acronym [SEP] snippet [SEP]" input format and the CLS token for the output. ``emb layer" and ``pred layer" stand for embedding layer and prediction layer respectively.}
\label{figure2}
\end{center}
\vskip -0.2in
\end{figure}

We carry out our investigation on objective alignment and FEL through concept-of-interest tagging (CT) and acronym detection (AD) tasks. The technical implementations of CT and AD are shown in Figure \ref{figure2}.

\textbf{Concept-of-interest tagging (CT)}. The goal of CT is to annotate the start and the end positions of concepts belonging to the categories of interest within a short query text. We define a concept as the shortest contiguous sequence of text that carries a standalone meaning. For example, query ``\textit{when was the first super mario released}" has four concepts ``\textit{when was}", ``\textit{the first}", ``\textit{super mario}" and ``\textit{released}". The concept categories of interest are determined by downstream applications. In this paper, we build a concept tagger to annotate all concepts except for the ones that overlap with the following six types of language features: (1) Five Ws (who, what, when, where, why); (2) Verbs and verb phrases; (3) Prepositions, conjunctions and determiners that are outside of fixed expressions; (4) Comparatives and superlatives; (5) Numbers and IDs; (6) Natural language-type book, song, movie titles. Based on these choices, our tagger should only annotate ``\textit{super mario}" in the example query above.

\textbf{Acronym detection (AD)}. The goal of AD is to determine whether a unigram is a valid acronym of any entity present in a text snippet. For example, given snippet ``\textit{much of world of warcraft's gameplay involves the completion of quests. these quests are usually available from npcs.}", unigram ``\textit{wow}" is a valid acronym as it matches to ``\textit{world of warcraft}". But unigrams ``\textit{tqa}" (these quests are) and ``\textit{woa}" are not valid acronyms because they either do not match to an entity or do not match any span of text. AD does not aim to memorize ``\textit{wow}" as the most common acronym for ``\textit{world of warcraft}", but only whether a unigram could be a valid acronym of an entity.

\subsection{Design Aligned Pretraining Objectives}
\label{subsectionalignedobjectives}

We designed aligned pretraining objectives Wikipedia hyperlink prediction for CT and pseudo acronym detection for AD.

\textbf{Wikipedia hyperlink prediction}. The majority of hyperlink texts on Wikipedia pages are named-entities, common nouns and noun phrases. They contain specialized concept understanding knowledge for performing CT. We extract the hyperlink metadata from the official Wikipedia dump and use the same CT 4-class classification to predict the start and the end of hyperlinks in our pretraining. It is noteworthy that training with Wikipedia hyperlinks does lead to annotations of the six unwanted categories described in Section \ref{subsectiontasks}. Many book, movie and song titles are natural language-like, starting with five Ws and containing verbs etc. Many product names and events contain numbers and IDs. We also found that, MLM alone can provide good concept understanding to the model, making the Wikipedia hyperlink prediction task partially redundant, we will provide more discussion on this in Section \ref{subsectionredundency}.

\textbf{Pseudo acronym detection}. To perform AD, the language model needs to understand both concepts and spelling. While MLM and Wikipedia hyperlink prediction inject concept understanding knowledge into the model, we use pseudo acronym detection task along with subword regularization \cite{Kudo2018} to teach the model spelling. Pseudo acronym detection is a binary classification task like AD, and we generate its training data in the following steps: (1) From a chunk of Wikipedia text, we take a random contiguous span of 2 to 6 unigrams. We concatenate their initials as a positive example; (2) For the same chunk of text we generate a negative example using one of two methods with equal probabilities - randomly mutating one to all of the letters in the positive unigram or taking the positive unigram belonging to another random chunk of text; (3) We make sure there is no accidental match for the negative examples.

\subsection{Pretraining}

Our pretraining dataset is composed of a filtered Wikipedia corpus (about 70\% English, 30\% non-English) and a set of internally collected web documents (about 50\% English, 50\% non-English). We created the model vocabulary from our web document corpus using the unigram language model algorithm in the sentencepiece library \cite{Sentencepiece}. Our vocabulary covers 111K unique tokens and about 6000 unique characters. The filtered Wikipedia corpus tokenizes to 4.4 billion tokens and the web document corpus tokenizes to 1.6 billion tokens.

We used four pretraining objectives: (1) MLM; (2) NSP; (3) Wikipedia hyperlink prediction and (4) Pseudo acronym detection. MLM and NSP are directly taken from the BERT paper \cite{Devlin2019}, but instead of masking tokens with equal probabilities, we mask rarer tokens with higher probabilities \cite{Conneau2019}. MLM and NSP labels are generated randomly on-the-fly and they are applied to both corpuses. Wikipedia hyperlink prediction and pseudo acronym detection labels are pre-generated across the Wikipedia corpus only. Subword regularization is applied to the web document corpus only to make pseudo acronym detection more challenging.

We trained a set of eleven language models that have the same architecture as BERT but differ in dimensions. Nine of them have sizes 256, 512, 768 embedding dimensions by 1, 2, 3 Transformer layers. The last two are both 768 by 3 models, one pretrained without Wikipedia hyperlink prediction, one pretrained without pseudo acronym detection. All models have 64-dimension Transformer heads and 3072 hidden dimensions. We use learning rate 0.0001 with Adam optimizer, dropout 0.1, batch size 256 and maximum sequence length 128. We train each model for 500 million examples, which is 10 epochs of the Wikipedia corpus (our web documents have shorter sequence lengths and higher examples to total token ratio). It takes two weeks to pretrain a 768 by 3 model on four Tesla P40 GPUs. Our model training code was forked from the XLM code base \cite{Conneau2019}.

\subsection{Finetuning}
\label{subsectionfinetuning}

The CT dataset contains 475 manually labeled English queries and 2199 token-level labels with label distribution 16\% S, 16\% E, 9\% S\&E, 59\% $\sim$(S\&E) (Figure \ref{figure2}). The 475 examples further split into a 275 example subset and a 200 example subset. The two subsets do not share any non-stopword unigrams. The AD dataset contains 1200 manually labeled English acronym-snippet pairs with label distribution 50\% 0 and 50\% 1. We created this dataset by taking 600 unique snippets each containing at least one unique named-entity with a known acronym. The known acronyms are the positive examples. We create one negative example for each snippet using one of the two methods with equal probabilities - generating a pseudo acronym detection positive example or randomly mutating one or more letters in the positive acronym unigram. We made sure that the acronym unigrams never appear in the snippet and there is no accidental false negatives in the negative examples.

For each data point presented in Section \ref{sectionresults}, we perform 5-fold cross validation with two random seeds each - a total of ten finetuning calculations. In the case of CT, instead of the standard 5-fold cross validation, we do five random samples from the 275 example subset to form the training sets and use the 200 example subset as the testset. By default, we use batch size 64, dropout 0.1, learning rate 0.0001, finetuning data size 200 and backpropagate only through the prediction layers. We also discuss the effects of different finetuning parameters in Section \ref{sectionresults}. Finetuning calculations typically finish within fifteen minutes on one Tesla P40 GPU.

\begin{table*}
\caption{\textbf{CT and AD accuracy/perplexity by model size.} Model accuracy and perplexity improve as model size increases. The boldface models statistically significantly outperform model 768-$\times$3*, which is pretrained without objective alignment (underlined). Models in row CT, 768- are pretrained without Wikipedia hyperlink prediction. Models in row AD, 768- are pretrained without pseudo acronym detection. Models in column 3* are finetuned after first discarding the prediction layers.}
\label{table1}
\vskip 0.15in
\begin{center}
\begin{small}
\begin{sc}
\begin{tabular}{clccccr}
\toprule
& & 1 & 2 & 3 & 3* \\
\midrule
\multirow{4}{*}{CT} & 256 & 77.8$\pm$0.5 / 1.75$\pm$0.01 & 80.0$\pm$0.7 / 1.70$\pm$0.04 & \textbf{81.9$\pm$0.6 / 1.58$\pm$0.01} & 79.6$\pm$0.7 / 1.70$\pm$0.04 \\
& 512 & 77.2$\pm$0.8 / 1.78$\pm$0.03 & \textbf{82.1$\pm$0.5 / 1.56$\pm$0.01} & \textbf{82.5$\pm$0.6 / 1.55$\pm$0.04} & \textbf{81.5$\pm$0.4 / 1.62$\pm$0.02} \\
& 768 & 79.0$\pm$0.7 / 1.71$\pm$0.01 & \textbf{83.7$\pm$0.3 / 1.53$\pm$0.01} & \textbf{83.9$\pm$0.3 / 1.50$\pm$0.01} & \textbf{82.4$\pm$0.8 / 1.58$\pm$0.02} \\
& 768- & / & / & / & \underline{79.1$\pm$0.9 / 1.76$\pm$0.06} \\
\midrule
\multirow{4}{*}{AD} & 256 & 56.9$\pm$0.6 / 1.98$\pm$0.01 & 61.6$\pm$0.3 / 1.92$\pm$0.02 & \textbf{65.3$\pm$1.0 / 1.86$\pm$0.02} & \textbf{65.4$\pm$0.9 / 1.86$\pm$0.02} \\
& 512 & 57.6$\pm$0.8 / 1.97$\pm$0.01 & \textbf{66.4$\pm$0.5 / 1.87$\pm$0.02} & \textbf{70.2$\pm$1.5 / 1.79$\pm$0.03} & \textbf{70.2$\pm$1.6 / 1.79$\pm$0.03} \\
& 768 & 59.8$\pm$0.8 / 1.99$\pm$0.03 & \textbf{70.7$\pm$1.7 / 1.81$\pm$0.05} & \textbf{73.8$\pm$0.8 / 1.79$\pm$0.04} & \textbf{73.8$\pm$0.5 / 1.80$\pm$0.04} \\
& 768- & / & / & / & \underline{63.9$\pm$1.3 / 1.96$\pm$0.05} \\
\bottomrule
\end{tabular}
\end{sc}
\end{small}
\end{center}
\vskip -0.1in
\end{table*}

\begin{table*}[t]
\caption{\textbf{CT and AD accuracy/perplexity by finetuning data size.} Model accuracy and perplexity improve as finetuning data size increases from 50 to 1000. The boldface models statistically significantly outperform model 768-$\times$3*, which is finetuned using 200 examples (underlined). Symbols ``-" and ``*" carry the same meanings as in Table \ref{table1}.}
\label{table2}
\vskip 0.15in
\begin{center}
\begin{small}
\begin{sc}
\begin{tabular}{clcccc}
\toprule
& & 768-$\times$3* & 512$\times$3* & 768$\times$3* & 768$\times$3  \\
\midrule
\multirow{4}{*}{CT} & 50 & 72.4$\pm$2.1 / 2.05$\pm$0.10 & 76.4$\pm$1.1 / 1.84$\pm$0.05 & 77.7$\pm$1.4 / 1.78$\pm$0.03 & \textbf{82.0$\pm$0.5 / 1.59$\pm$0.02}\\
& 100 & 76.5$\pm$1.3 / 1.88$\pm$0.11 & 79.1$\pm$1.0 / 1.74$\pm$0.06 & 80.3$\pm$0.9 / 1.65$\pm$0.02 & \textbf{83.7$\pm$0.7 / 1.52$\pm$0.01}\\
& 150 & 77.9$\pm$1.0 / 1.81$\pm$0.10 & 80.4$\pm$0.8 / 1.69$\pm$0.07 & \textbf{81.8$\pm$0.5 / 1.60$\pm$0.02} & \textbf{83.7$\pm$0.7 / 1.51$\pm$0.01}\\
& 200 & \underline{79.1$\pm$0.9 / 1.76$\pm$0.06} & \textbf{81.5$\pm$0.4 / 1.62$\pm$0.02} & \textbf{82.4$\pm$0.8 / 1.58$\pm$0.02} & \textbf{83.9$\pm$0.3 / 1.50$\pm$0.01}\\
\midrule
\multirow{5}{*}{AD} & 50 & 58.7$\pm$1.8 / 2.00$\pm$0.05 & 64.3$\pm$0.6 / 1.91$\pm$0.04 & 65.2$\pm$1.4 / 1.97$\pm$0.08 & 65.0$\pm$1.6 / 1.98$\pm$0.05\\
& 100 & 61.0$\pm$0.6 / 1.99$\pm$0.05 & \textbf{66.3$\pm$1.0 / 1.87$\pm$0.03} & \textbf{70.3$\pm$0.6 / 1.92$\pm$0.06} & \textbf{70.1$\pm$0.8 / 1.89$\pm$0.06}\\
& 150 & 62.4$\pm$1.8 / 1.97$\pm$0.07 & \textbf{68.6$\pm$1.7 / 1.83$\pm$0.03} & \textbf{72.5$\pm$0.5 / 1.87$\pm$0.04} & \textbf{72.3$\pm$0.7 / 1.83$\pm$0.03}\\
& 200 & \underline{63.9$\pm$1.3 / 1.96$\pm$0.05} & \textbf{70.2$\pm$1.6 / 1.79$\pm$0.03} & \textbf{73.8$\pm$0.5 / 1.80$\pm$0.04} & \textbf{73.8$\pm$0.8 / 1.79$\pm$0.04}\\
& 1000 & \textbf{72.9$\pm$2.5 / 1.78$\pm$0.05} & \textbf{77.2$\pm$3.1 / 1.67$\pm$0.06} & \textbf{84.7$\pm$1.3 / 1.49$\pm$0.04} & \textbf{85.5$\pm$1.4 / 1.47$\pm$0.03}\\
\bottomrule
\end{tabular}
\end{sc}
\end{small}
\end{center}
\vskip -0.1in
\end{table*}

\section{Results}
\label{sectionresults}

We evaluate the accuracy and perplexity of CT and AD tasks across the eleven language models with different finetuning parameters. All values are averaged over ten finetuning calculations (Section \ref{subsectionfinetuning}), and we provide the one standard deviation values in Tables \ref{table1}, \ref{table2} and \ref{table4}. We use the standard accuracy and perplexity calculation formulas, the exact computation methods can be found in the XLM code base \cite{Conneau2019}. Accuracy, i.e. precision without thresholding, is the determining metric for our NLU applications, but we also use perplexity to gauge the overall quality of our models.

\vspace{5mm}
\subsection{Case Studies}

\vspace{3.5mm}
{\fontsize{8}{8}\selectfont
\begin{verbatim}
WHP: [when was [the first [super mario] released]
 CT:  when was  the first [super mario] released

WHP: [skyrim] [100% completion]
 CT: [skyrim]  100% completion
\end{verbatim}}

Wikipedia hyperlink prediction (WHP) and CT lead to different outcomes of query tagging. We use symbols ``[" and ``]" to denote the predicted concept boundaries by the model. Concepts containing five Ws (``\textit{when}"), verbs (``\textit{was}", ``\textit{released}") and numbers (``\textit{first}", ``\textit{100\%}") are not annotated in CT. Note that Wikipedia hyperlink prediction leads to nested annotations even though hyperlinks are non-overlapping. This is because the classification loss of a token is conditioned independently of the classes of their context tokens.

{\fontsize{8}{8}\selectfont
\begin{verbatim}
SNIPPET:
    much of world of warcraft's gameplay involves
    the completion of quests. these quests are
    usually available from npcs.

ACRONYMS:
    wow: PAD score = 0.80, AD score = 0.71
    tqa: PAD score = 0.95, AD score = 0.34
    woa: PAD score = 0.09, AD score = 0.18
\end{verbatim}}

Pseudo acronym detection (PAD) gives ``\textit{wow}" and ``\textit{tqa}" positive labels (score$>$0.5), while AD only gives ``\textit{wow}" a positive label. This is because ``\textit{wow}" is the only unigram that matches to an entity. The model scores are the softmax probabilities of the positive class. All examples above are generated by one of the finetuned 768 by 3 models.

\subsection{Objective Alignment and Model Size}
\label{subsectionmodelsize}

Since the models pretrained without Wikipedia hyperlink prediction and pseudo acronym detection (denoted by ``-") do not have pre-minimized prediction layers for CT and AD, we discard the prediction layers of the pretrained models for fair comparisons (denoted by ``*"). Table \ref{table1} shows that smaller language models pretrained with objective alignment can outperform bigger language models pretrained without objective alignment. Model 768$\times$3* achieves +3.3\% accuracy in CT and +9.9\% accuracy in AD compared to model 768-$\times$3* of the same size. When comparing across model sizes, the smallest models that statistically significantly outperform model 768-$\times$3* are model 512$\times$3* and model 256$\times$3* for CT (+1.4\%) and AD  (+1.5\%) respectively.

Keeping the pre-minimized prediction layers of the objective aligned models further improves the accuracy of the finetuning results of CT but has little impacts on AD. On the other hand, AD gets bigger accuracy boost than CT from both objective alignment and model size increasement. This is most likely because of the standard language model tokenization algorithm we used. The algorithm tokenizes words into the longest most probable subwords \cite{Wu2016, Kudo2018} and as a result, the model does not know the letter-by-letter spelling of a word explicitly. This makes AD a more challenging task for the language models we trained.

As expected, the best performing model of both CT and AD tasks is the largest model we trained, model 768$\times$3. In terms of the embedding dimension, CT accuracy increased 2.0\% and AD accuracy increased 8.5\% from model 256$\times$3 to model 768$\times$3. In terms of the number of Transformer layers, CT accuracy increased 4.9\% and AD accuracy increased 14.0\% from model 768$\times$1 to model 768$\times$3. Model 256$\times$3 significantly outperforms model 768$\times$1, which shows, for the vanilla Transformer architecture, at least two Transformer layers are recommended to learn tasks of moderate complexity.

\subsection{FEL and Training Data Size}

Table \ref{table2} shows that objective alignment reduces the amount of finetuning examples required to reach a targeted accuracy, and good objective alignment can reduce the minimum number of finetuning examples required to the order of hundreds. For example, for the AD task, model 768$\times$3* can reach accuracy of 72.5\% using 150 finetuning examples while model 768-$\times$3* requires 1000 examples to reach the same accuracy.

The accuracy of model 768$\times$3 improves the most from using 50 examples to 100 examples for both CT and AD. CT accuracy stops improving after 100 examples, while AD accuracy keeps improving linearly from 100 to 1000 examples. This again shows that AD is a more challenging task and requires more examples to learn. Table \ref{table2} also shows that the observations made in Section \ref{subsectionmodelsize} hold across all FEL training data sizes - the accuracy of model 768$\times$3 $>$ model 768$\times$3* $>$ model 512$\times$3* $>$ model 768-$\times$3*.

\subsection{FEL Training Parameters}
The Transformer language models we use here are composed of the embedding layer, the Transformer layers and the prediction layers (Figure \ref{figure2}). We perform finetuning by backpropagating through (1) only the prediction layers, (2) the prediction and the Transformer layers, and (3) all of the layers (Table \ref{table3}).

When backpropagating through only the prediction layers, the model reaches the best accuracy and perplexity at the same time. But when we extend the backpropagation to the other layers, the model reaches the best perplexity first, then reaches the best accuracy while perplexity continues to increase. We see this as a sign of overfitting - the accuracy of a portion of the testset examples increases while the accuracy of the rest of the testset worsens. Even though accuracy is generally the determining metric for our applications, we pick the model with the best perplexity given that the corresponding accuracy is acceptable. CT-AD multitasking while backpropagating through the prediction and transformer layers achieves the best accuracy and perplexity for both tasks, showing that multitasking significantly benefits FEL in its low-resource regime. Despite this, we choose to backpropagate only through the prediction layers as our best practice, because it best prevents overfitting.

We also experimented with the choice of learning rate for FEL (Table \ref{table4}). We found that when backpropagating through the prediction layers only, the performance of CT and AD tasks are not very sensitive to learning rates in the range of 0.001 and 0.00001. AD shows accuracy loss and perplexity gain as learning rate increases but no significant combined difference. We speculate that the loss functions for CT and AD are smooth near their minima or that they have multiple equally good closely spaced local minima. We chose a learning rate of 0.0001 for the other FEL calculations in this paper, as it provides the best combination of performance and training time.

{\setlength{\tabcolsep}{4pt}
\begin{table}
\caption{\textbf{FEL backpropagation scheme.} Backpropagating through both prediction (PRED) and Transformer (TRM) layers while multitasking (MT) achieves the best accuracy/perplexity combination. The best-accuracy models are overfitted when backpropagating through the Transformer and embedding (EMB) layers. We recommend selecting models using perplexity. Results are generated using model 768$\times$3.}
\label{table3}
\vskip 0.15in
\begin{center}
\begin{small}
\begin{sc}
\begin{tabular}{clcc}
\toprule
& backpropagation & acc/best-ppl & best-acc/ppl\\
\midrule
\multirow{4}{*}{CT} & pred & 83.8 / 1.50 & 83.9 / 1.50\\
& mt pred+trm & 84.8 / 1.47 & 86.0 / 1.75\\
& pred+trm & 85.0 / 1.47 & 85.9 / 1.63\\
& pred+trm+emb & 84.8 / 1.48 & 85.9 / 1.67\\
\midrule
\multirow{4}{*}{AD} & pred & 72.3 / 1.75 & 73.8 / 1.79\\
& mt pred+trm & 74.2 / 1.69 & 84.8 / 2.38\\
& pred+trm & 73.8 / 1.70 & 83.8 / 2.65\\
& pred+trm+emb & 73.3 / 1.71 & 83.9 / 2.63\\
\bottomrule
\end{tabular}
\end{sc}
\end{small}
\end{center}
\vskip -0.1in
\end{table}
}

\begin{table}[t]
\caption{\textbf{FEL learning rate.} Learning rate does not have a significant impact on the finetuning accuracy and perplexity of CT and AD. Smaller learning rates require more epochs of training. Results are generated using model 768$\times$3.}
\label{table4}
\vskip 0.15in
\begin{center}
\begin{small}
\begin{sc}
\begin{tabular}{clcc}
\toprule
& & acc/ppl & epochs \\
\midrule
\multirow{3}{*}{CT} & 0.001 & 83.7 / 1.53 & 4.6$\pm$1.6\\
& 0.0001 & 83.9 / 1.50 & 33.1$\pm$6.9\\
& 0.00001 & 83.8 / 1.50 & 291.4$\pm$40.4\\
\midrule
\multirow{3}{*}{AD} & 0.001 & 74.8 / 2.01 & 84.4$\pm$17.9\\
& 0.0001 & 73.8 / 1.79 & 156.2$\pm$33.3\\
& 0.00001 & 71.7 / 1.76 & 903.5$\pm$128.4\\
\bottomrule
\end{tabular}
\end{sc}
\end{small}
\end{center}
\vskip -0.1in
\end{table}

\begin{table}[t]
\caption{\textbf{Objective alignment vs. objective diversity.} Pseudo acronym detection provides the language model more unique information than Wikipedia hyperlink prediction. It not only significantly improves the AD accuracy, but also improves the CT accuracy by more than 1\%. Symbol ``-" carries the same meaning as in Table \ref{table1}.}
\label{table5}
\vskip 0.15in
\begin{center}
\begin{small}
\begin{sc}
\begin{tabular}{cccc}
\toprule
& 768$\times$3* & CT 768-$\times$3* & AD 768-$\times$3* \\
\midrule
CT & 82.4 / 1.58 & 79.1 /1.76 & 80.6 / 1.65 \\
AD & 73.8	/ 1.80 & 73.9 / 1.72 & 63.9 / 1.96 \\
\bottomrule
\end{tabular}
\end{sc}
\end{small}
\end{center}
\vskip -0.1in
\end{table}

\begin{table}[t]
\caption{\textbf{MLM and NSP accuracy/perplexity.} The MLM and NSP accuracy and perplexity remain the same with and without the aligned pretraining objectives. Symbol ``-" carries the same meaning as in Table \ref{table1}.}
\label{table6}
\vskip 0.15in
\begin{center}
\begin{small}
\begin{sc}
\begin{tabular}{cccc}
\toprule
& 768$\times$3 & CT 768-$\times$3 & AD 768-$\times$3 \\
\midrule
MLM & 52.4 / 14.1 & 52.6 / 14.0 & 52.8 / 13.5 \\
NSP & 99.1 / 1.02	 & 99.1 / 1.02 & 99.2 / 1.02 \\
\bottomrule
\end{tabular}
\end{sc}
\end{small}
\end{center}
\vskip -0.1in
\end{table}

\section{Discussion}

\subsection{Pretraining Objective Diversity}
\label{subsectionredundency}

In Table \ref{table1}, model 768$\times$3* outperforms model 768-$\times$3* by +3.3\% accuracy in CT and +9.9\% accuracy in AD. The improvement of the CT task from objective alignment is 3 times less than that of the AD task. We think this is because that MLM and Wikipedia hyperlink prediction are intrinsically similar. The model has acquired most of the knowledge needed to perform CT from MLM and Wikipedia hyperlink prediction does not add significant amount of new information. Unlike Wikipedia hyperlink prediction, pseudo acronym detection resembles neither MLM nor NSP. It teaches unique new knowledge to the model that is necessary to perform AD. Additionally, in Table \ref{table5}, model CT 768-$\times$3* pretrained without Wikipedia hyperlink prediction gives the same AD accuracy as model 768$\times$3* pretrained with all four pretraining objectives. Model AD 768-$\times$3* pretrained without pseudo acronym detection instead reaches 1.8\% less CT accuracy compared to model 768$\times$3*. This means, Wikipedia hyperlink prediction and pseudo acronym detection are approximately responsible for 1.5\% and 1.8\% of the 3.3\% CT accuracy improvement. On the other hand, pseudo acronym detection is responsible for the entire 9.9\% of the accuracy improvement for AD. These results show that pseudo acronym detection not only helps the model perform AD better but also has the potential to benefit other tasks that are seemingly less related. In future works, we plan to build a more universal language model by designing more diverse pretraining objectives to align better with a broader range of finetuning tasks.

\subsection{Low-Resource Tasks vs. High-Resource Tasks}

Table \ref{table6} shows that our added pretraining objectives do not lead to significant variations of the accuracy and perplexity of MLM and NSP. Not surprisingly, while objective alignment and objective diversity could improve the performance of low-resource tasks, their effects on high-resource tasks are minimal. Models 768-$\times$3 (CT) and 768-$\times$3 (AD) show slightly better accuracy and perplexity than model 768$\times$3. This is mostly likely because that we pretrain all models for 500 million examples regardless of the choice of pretraining tasks. When training without Wikipedia hyperlink prediction or pseudo acronym detection, the model effectively see more MLM and NSP examples.

\subsection{The Limits of FEL}
\label{subsectionfelapplies}

Objective alignment is a technique that applies to all finetuning tasks, but the same cannot be said for FEL. We apply FEL to CT and AD because they only require the understanding of a few rules that are made of the fundamental properties of a language - concept-boundary, part of speech and spelling. We show that these small numbers of rules can be explained by just a few examples. In fact, AD is pushing the limit of ``developer-affordable" FEL. Even though model 768$\times$3 can reach a reasonable AD accuracy using 200 finetuning examples, for the best performance, 1000 examples or more are recommended, which is still a very affordable size when hiring human judges. FEL does not apply to tasks that require heavy memorization. For example, a task to tag only the names of celebrities or a task to predict the likelihood of a unigram being the official acronym of a named-entity.

\section{Conclusion}

We developed the language model training techniques objective alignment and FEL to tackle two challenges in building NLU applications: (1) Speeding up language model inference and (2) Shortage of finetuning data for application-specific tasks. We produced a high-accuracy concept-of-interest tagger and a medium-accuracy acronym detector using our techniques. Both models have dimension 768 by 3 and were trained with 200 finetuning examples each. This small model size meets our real-time inference latency requirements, and the small finetuning dataset can be generated by the developers themselves or a few human judges. Overall, our language model training solution greatly improves the efficiency and agility of our NLU application development cycles.

\bibliography{fel_paper_bib}

\begin{thebibliography}{26}
\providecommand{\natexlab}[1]{#1}
\providecommand{\url}[1]{\texttt{#1}}
\expandafter\ifx\csname urlstyle\endcsname\relax
  \providecommand{\doi}[1]{doi: #1}\else
  \providecommand{\doi}{doi: \begingroup \urlstyle{rm}\Url}\fi

\bibitem[Conneau \& Lample(2019)Conneau and Lample]{Conneau2019}
Conneau, A. and Lample, G.
\newblock {Cross-lingual Language Model Pretraining}.
\newblock In \emph{Advances in Neural Information Processing Systems 32}, pp.\
  7057--7067. Curran Associates, Inc., 2019.

\bibitem[Conneau et~al.(2019)Conneau, Khandelwal, Goyal, Chaudhary, Wenzek,
  Guzmán, Grave, Ott, Zettlemoyer, and Stoyanov]{Conneau20192}
Conneau, A., Khandelwal, K., Goyal, N., Chaudhary, V., Wenzek, G., Guzmán, F.,
  Grave, E., Ott, M., Zettlemoyer, L., and Stoyanov, V.
\newblock {Unsupervised Cross-lingual Representation Learning at Scale}.
\newblock \emph{arXiv preprint arXiv:1911.02116}, 2019.

\bibitem[Dai et~al.(2019)Dai, Yang, Yang, Carbonell, Le, and
  Salakhutdinov]{Dai2019}
Dai, Z., Yang, Z., Yang, Y., Carbonell, J., Le, Q.~V., and Salakhutdinov, R.
\newblock {Transformer-XL: Attentive Language Models Beyond a Fixed-Length
  Context}.
\newblock \emph{arXiv preprint arXiv:1901.02860}, 2019.

\bibitem[Devlin et~al.(2019)Devlin, Chang, Lee, and Toutanova]{Devlin2019}
Devlin, J., Chang, M.-W., Lee, K., and Toutanova, K.
\newblock {BERT: Pre-training of Deep Bidirectional Transformers for Language
  Understanding}.
\newblock In \emph{Proc. of NAACL}, 2019.

\bibitem[Dong et~al.(2019)Dong, Yang, Wang, Wei, Liu, Wang, Gao, Zhou, and
  Hon]{Dong2019}
Dong, L., Yang, N., Wang, W., Wei, F., Liu, X., Wang, Y., Gao, J., Zhou, M.,
  and Hon, H.-W.
\newblock {Unified Language Model Pre-training for Natural Language
  Understanding and Generation}.
\newblock In \emph{Advances in Neural Information Processing Systems 32}, pp.\
  13042--13054. Curran Associates, Inc., 2019.

\bibitem[Hinton et~al.(2015)Hinton, Vinyals, and Dean]{Hinton2015}
Hinton, G., Vinyals, O., and Dean, J.
\newblock {Distilling the Knowledge in a Neural Network}.
\newblock In \emph{NIPS Deep Learning and Representation Learning Workshop},
  2015.

\bibitem[Jiao et~al.(2019)Jiao, Yin, Shang, Jiang, Chen, Li, Wang, and
  Liu]{Jiao2019}
Jiao, X., Yin, Y., Shang, L., Jiang, X., Chen, X., Li, L., Wang, F., and Liu,
  Q.
\newblock {TinyBERT: Distilling BERT for Natural Language Understanding}.
\newblock \emph{arXiv preprint arXiv:1909.10351}, 2019.

\bibitem[Junczys-Dowmunt et~al.(2018)Junczys-Dowmunt, Grundkiewicz, Dwojak,
  Hoang, Heafield, Neckermann, Seide, Germann, Aji, Bogoychev, Martins, and
  Birch]{Junczysdowmunt2018}
Junczys-Dowmunt, M., Grundkiewicz, R., Dwojak, T., Hoang, H., Heafield, K.,
  Neckermann, T., Seide, F., Germann, U., Aji, A.~F., Bogoychev, N., Martins,
  A. F.~T., and Birch, A.
\newblock {Marian: Fast Neural Machine Translation in C++}.
\newblock In \emph{Proceedings of ACL 2018, System Demonstrations}, pp.\
  116--121, 2018.

\bibitem[Kudo(2018{\natexlab{a}})]{Kudo2018}
Kudo, T.
\newblock {Subword Regularization: Improving Neural Network Translation Models
  with Multiple Subword Candidates}.
\newblock In \emph{Proceedings of the 56th Annual Meeting of the Association
  for Computational Linguistics}, pp.\  66--75. 2018{\natexlab{a}}.

\bibitem[Kudo(2018{\natexlab{b}})]{Sentencepiece}
Kudo, T.
\newblock {sentencepiece}.
\newblock \url{https://github.com/google/sentencepiece}, 2018{\natexlab{b}}.
\newblock Accessed: July 2019.

\bibitem[Lan et~al.(2020)Lan, Chen, Goodman, Gimpel, Sharma, and
  Soricut]{Lan2020}
Lan, Z., Chen, M., Goodman, S., Gimpel, K., Sharma, P., and Soricut, R.
\newblock {ALBERT: A Lite BERT for Self-supervised Learning of Language
  Representations}.
\newblock In \emph{International Conference on Learning Representations}, 2020.

\bibitem[Lin et~al.(2018)Lin, Yang, Stoyanov, and Ji]{Lin2018}
Lin, Y., Yang, S., Stoyanov, V., and Ji, H.
\newblock {A Multi-lingual Multi-task Architecture for Low-resource Sequence
  Labeling}.
\newblock In \emph{Proceedings of the 56th Annual Meeting of the Association
  for Computational Linguistics (Volume 1: Long Papers)}, pp.\  799--809, 2018.

\bibitem[Liu et~al.(2019{\natexlab{a}})Liu, He, Chen, and Gao]{Liu2018}
Liu, X., He, P., Chen, W., and Gao, J.
\newblock {Multi-Task Deep Neural Networks for Natural Language Understanding}.
\newblock In \emph{{Proceedings of the 57th Annual Meeting of the Association
  for Computational Linguistics}}, pp.\  4487--4496, 2019{\natexlab{a}}.

\bibitem[Liu et~al.(2019{\natexlab{b}})Liu, He, Chen, and Gao]{Liu22019}
Liu, X., He, P., Chen, W., and Gao, J.
\newblock {Improving Multi-Task Deep Neural Networks via Knowledge Distillation
  for Natural Language Understanding}.
\newblock \emph{arXiv preprint arXiv:1904.09482}, 2019{\natexlab{b}}.

\bibitem[Liu et~al.(2019{\natexlab{c}})Liu, Ott, Goyal, Du, Joshi, Chen, Levy,
  Lewis, Zettlemoyer, and Stoyanov]{Liu2019}
Liu, Y., Ott, M., Goyal, N., Du, J., Joshi, M., Chen, D., Levy, O., Lewis, M.,
  Zettlemoyer, L., and Stoyanov, V.
\newblock {RoBERTa: A Robustly Optimized BERT Pretraining Approach}.
\newblock \emph{arXiv preprint arXiv:1907.11692}, 2019{\natexlab{c}}.

\bibitem[Peters et~al.(2018)Peters, Neumann, Iyyer, Gardner, Clark, Lee, and
  Zettlemoyer]{Peters2018}
Peters, M.~E., Neumann, M., Iyyer, M., Gardner, M., Clark, C., Lee, K., and
  Zettlemoyer, L.
\newblock {Deep Contextualized Word Representations}.
\newblock In \emph{Proc. of NAACL}, 2018.

\bibitem[Radford et~al.(2018)Radford, Narasimhan, Salimans, and
  Sutskever]{Radford2018}
Radford, A., Narasimhan, K., Salimans, T., and Sutskever, I.
\newblock {Improving Language Understanding by Generative Pre-Training}.
\newblock Technical report, OpenAI, 2018.

\bibitem[Radford et~al.(2019)Radford, Wu, Child, Luan, Amodei, and
  Sutskever]{Radford2019}
Radford, A., Wu, J., Child, R., Luan, D., Amodei, D., and Sutskever, I.
\newblock {Language Models are Unsupervised Multitask Learners}.
\newblock Technical report, OpenAI, 2019.

\bibitem[Raffel et~al.(2019)Raffel, Shazeer, Roberts, Lee, Narang, Matena,
  Zhou, Li, and Liu]{Raffel2019}
Raffel, C., Shazeer, N., Roberts, A., Lee, K., Narang, S., Matena, M., Zhou,
  Y., Li, W., and Liu, P.~J.
\newblock {Exploring the Limits of Transfer Learning with a Unified
  Text-to-Text Transformer}.
\newblock \emph{arXiv preprint arXiv:1910.10683}, 2019.

\bibitem[Rajpurkar et~al.(2016)Rajpurkar, Zhang, Lopyrev, and
  Liang]{Rajpurkar2016}
Rajpurkar, P., Zhang, J., Lopyrev, K., and Liang, P.
\newblock {SQuAD: 100,000+ Questions for Machine Comprehension of Text}.
\newblock In \emph{Proceedings of the 2016 Conference on Empirical Methods in
  Natural Language Processing}, pp.\  2383--2392, 2016.

\bibitem[Sun et~al.(2020)Sun, Wang, Li, Feng, Tian, Wu, and Wang]{Sun2020}
Sun, Y., Wang, S., Li, Y., Feng, S., Tian, H., Wu, H., and Wang, H.
\newblock {ERNIE 2.0: A Continual Pre-training Framework for Language
  Understanding}.
\newblock \emph{AAAI}, 2020.

\bibitem[Vaswani et~al.(2017)Vaswani, Shazeer, Parmar, Uszkoreit, Jones, Gomez,
  Kaiser, and Polosukhin]{Vaswani2017}
Vaswani, A., Shazeer, N., Parmar, N., Uszkoreit, J., Jones, L., Gomez, A.~N.,
  Kaiser, {\L}., and Polosukhin, I.
\newblock {Attention is All You Need}.
\newblock In \emph{Advances in Neural Information Processing Systems 30}, pp.\
  5998--6008. 2017.

\bibitem[Wang et~al.(2019)Wang, Singh, Michael, Hill, Levy, and
  Bowman]{Wang2019}
Wang, A., Singh, A., Michael, J., Hill, F., Levy, O., and Bowman, S.~R.
\newblock {GLUE: A Multi-Task Benchmark and Analysis Platform for Natural
  Language Understanding}.
\newblock In \emph{ICLR}, 2019.

\bibitem[Wu et~al.(2016)Wu, Schuster, Chen, Le, Norouzi, Macherey, Krikun, Cao,
  Gao, Macherey, Klingner, Shah, Johnson, Liu, Łukasz Kaiser, Gouws, Kato,
  Kudo, Kazawa, Stevens, Kurian, Patil, Wang, Young, Smith, Riesa, Rudnick,
  Vinyals, Corrado, Hughes, and Dean]{Wu2016}
Wu, Y., Schuster, M., Chen, Z., Le, Q.~V., Norouzi, M., Macherey, W., Krikun,
  M., Cao, Y., Gao, Q., Macherey, K., Klingner, J., Shah, A., Johnson, M., Liu,
  X., Łukasz Kaiser, Gouws, S., Kato, Y., Kudo, T., Kazawa, H., Stevens, K.,
  Kurian, G., Patil, N., Wang, W., Young, C., Smith, J., Riesa, J., Rudnick,
  A., Vinyals, O., Corrado, G., Hughes, M., and Dean, J.
\newblock {Google’s Neural Machine Translation System: Bridging the Gap
  Between Human and Machine Translation}.
\newblock \emph{arXiv preprint arXiv:1609.08144}, 2016.

\bibitem[Yang et~al.(2019)Yang, Dai, Yang, Carbonell, Salakhutdinov, and
  Le]{Yang2019}
Yang, Z., Dai, Z., Yang, Y., Carbonell, J., Salakhutdinov, R.~R., and Le, Q.~V.
\newblock {XLNet: Generalized Autoregressive Pretraining for Language
  Understanding}.
\newblock In \emph{Advances in Neural Information Processing Systems 32}, pp.\
  5754--5764. 2019.

\bibitem[Zhang et~al.(2018)Zhang, Rajbhandari, Wang, and He]{Zhang2018}
Zhang, M., Rajbhandari, S., Wang, W., and He, Y.
\newblock {DeepCPU: Serving RNN-based Deep Learning Models 10x Faster}.
\newblock In \emph{{USENIX ATC 18}}, pp.\  951--965, 2018.

\end{thebibliography}
\bibliographystyle{icml2020}

\end{document}